\documentclass[10pt,twocolumn,letterpaper]{article}

\usepackage[accsupp]{axessibility}  

\usepackage{cvpr}              
\usepackage{multirow}
\usepackage{svg}
\usepackage{float}

\definecolor{cvprblue}{rgb}{0.21,0.49,0.74}
\usepackage[pagebackref,breaklinks,colorlinks,allcolors=cvprblue]{hyperref}

\def\paperID{*****} 
\def\confName{CVPR}
\def\confYear{2026}

\title{UniSemAlign: Text–Prototype Alignment with a Foundation Encoder for Semi-Supervised Histopathology Segmentation}

\usepackage[T1]{fontenc}
\usepackage{babel}

\author{
    Le-Van Thai$^{1,*}$ \quad 
    Tien Dat Nguyen$^{1,*}$ \quad 
    Hoai Nhan Pham$^{1}$ \quad 
    Lan Anh Dinh Thi$^{2}$ \\[0.3cm]
    Duy-Dong Nguyen$^{1}$ \quad 
    Ngoc Lam Quang Bui$^{1}$ \\[0.5cm]
    $^{1}$AI VIETNAM Lab, Vietnam \\
    $^{2}$Hanoi University of Science and Technology, Vietnam \\
}

\begin{document}
\maketitle
\def\thefootnote{*}\footnotetext{Equal contribution}
\def\thefootnote{\arabic{footnote}}
\begin{abstract}
Semi-supervised semantic segmentation in computational pathology remains challenging due to scarce pixel-level annotations and unreliable pseudo-label supervision. We propose UniSemAlign, a dual-modal semantic alignment framework that enhances visual segmentation by injecting explicit class-level structure into pixel-wise learning. Built upon a pathology-pretrained Transformer encoder, UniSemAlign introduces complementary prototype-level and text-level alignment branches in a shared embedding space, providing structured guidance that reduces class ambiguity and stabilizes pseudo-label refinement. The aligned representations are fused with visual predictions to generate more reliable supervision for unlabeled histopathology images. The framework is trained end-to-end with supervised segmentation, cross-view consistency, and cross-modal alignment objectives. Extensive experiments on the GlaS and CRAG datasets demonstrate that UniSemAlign substantially outperforms recent semi-supervised baselines under limited supervision, achieving Dice improvements of up to 2.6\% on GlaS and 8.6\% on CRAG with only 10\% labeled data, and strong improvements at 20\% supervision. Code is available at: 
\href{https://github.com/thailevann/UniSemAlign}
{https://github.com/thailevann/UniSemAlign}

\end{abstract}    

\section{Introduction}
\label{sec:intro}

Recent advances in deep learning for medical imaging have highlighted tissue semantic segmentation as essential for automated pathology analysis \cite{hashimoto2020multi, zhang2022pixelseg} and clinical decision-making \cite{frank2023accurate, li2023scribblevc}. Fully-supervised medical image segmentation methods, ranging from foundational CNNs \cite{long2015fully, ronneberger2015u, chen2018encoderdecoderatrousseparableconvolution} to advanced Transformer-based architectures \cite{chen2021transunet, cao2022swin}, have achieved remarkable success in biomedical segmentation by effectively capturing multi-scale context and complex feature dependencies. Despite their efficacy, fully-supervised models require extensive pixel-level annotations, which are both labor-intensive and dependent on specialized clinical expertise \cite{wang2021annotation, shen2023co, zhang2024dslsm}. The challenge is particularly acute in histopathology, where complex morphological variations and high intra-class heterogeneity often result in ambiguous tissue boundaries, making precise and consistent annotation extremely difficult \cite{kapse2024si, li2024boundary, wu2025learning}.

Semi-supervised learning (SSL) addresses this challenge by using abundant unlabeled data with limited labeled samples \cite{han2024deep, qin2024urca, nadeem2025segmenting}. Common SSL segmentation methods, such as consistency regularization \cite{sohn2020fixmatch, yang2023unimatch} and pseudo-labeling \cite{bai2023bidirectional}, often struggle with low-quality pseudo-labels that introduce errors and reduce training stability \cite{xie2022clims, he2025trustmatch}, particularly in histopathology, where overlapping tissues and morphological ambiguity are prevalent \cite{thompson2022pseudo, wang2022ssa}. This issue stems from limited semantic guidance and underutilization of domain-specific pathological knowledge, which hinders fine-grained distinction \cite{vu2025semi, wang2022multi}.

Vision-language models (VLMs) and pathology-specific foundation models present richer semantic priors through image-text alignment \cite{radford2021learning, wang2022medclip, wang2022multi, yi2023simple}. Domain-adapted models such as PLIP \cite{huang2023visual} and CONCH \cite{lu2024visual}, as well as general-purpose models like UNI \cite{chen2024towards}, further enhance visual representations through image-text alignment and large-scale pretraining \cite{zhou2024knowledge}. Although these models are effective for global understanding and zero/few-shot tasks, their direct application to semi-supervised dense prediction in histopathology remains limited by domain shift, sensitivity to distribution variations \cite{gilal2025pathvlm, sikaroudi2023generalization}, and reliance on static prompts that fail to capture the subtle morphological nuances of histopathology \cite{fu2025multimodal, xu2024semi}. As a result, current VLM-based methods often fail to provide sufficient flexible and precise pixel-level semantic supervision to fully address the challenges of limited annotations in computational pathology \cite{pan2024dusss}.

Motivated by these limitations, we propose UniSemAlign, a dual-modal semantic alignment framework for semi-supervised histopathology segmentation. The proposed approach aligns visual representations from the UNI encoder \cite{chen2024towards} with textual semantics derived from a frozen CONCH text encoder \cite{lu2024visual} using CoOp-style prompt learning \cite{zhou2022learning}. Specifically, class-specific prototypes and prompt-guided text embeddings are used to provide complementary semantic cues for pixel-level prediction. The resulting alignment signals are integrated with the segmentation decoder to guide pseudo-label generation and refinement. Combined with a CorrMatch-style weak-to-strong consistency framework \cite{sun2023corrmatchlabelpropagationcorrelation}, UniSemAlign effectively leverages foundation model priors to improve pseudo-label reliability and segmentation performance under limited annotations.

Our main contributions are as follows:
\begin{itemize}

    \item We introduce a dual-modal semantic alignment framework for semi-supervised histopathology segmentation. This framework integrates learnable class prototypes and text-driven representations, delivering explicit class-level semantic guidance for pixel-wise learning and pseudo-label refinement.

    \item We leverage UNI as the encoder in a semi-supervised setting, demonstrating that domain-specific large-scale pretraining can be effectively combined with cross-modal semantic alignment to enhance representation robustness under limited annotations.

    \item Our method achieves superior performance on the GlaS and CRAG benchmarks, substantially narrowing the gap between semi-supervised and fully supervised segmentation in computational pathology.
\end{itemize}


\section{Related Works}
\label{sec:related_work}

\subsection{Semi-Supervised Semantic Segmentation}
Semantic segmentation demands precise pixel-level accuracy, making manual annotation highly time-consuming \cite{ronneberger2015u, long2015fully}. Semi-supervised frameworks mitigate this challenge by introducing semi-supervised learning (SSL) into spatial prediction tasks \cite{zhu2005semi, han2024deep}. Early approaches, including Cross Pseudo Supervision (CPS) \cite{chen2021cps} and Cross-Consistency Training (CCT) \cite{ouali2020semi}, promote model agreement on predictions between either two networks or multiple decoder heads within a single network. Concurrently, augmentation-based techniques such as CutMix-Seg \cite{french2020semi} and ST++ \cite{yang2023unimatch} improve generalization by generating composite images from patches and enforcing consistency in predictions across these artificial samples.

Despite these advancements, traditional self-training approaches remain highly susceptible to confirmation bias, which causes models to overfit to inaccurate pseudo-masks, particularly at ambiguous semantic boundaries \cite{tarvainen2017mean, sohn2020fixmatch}. Recent state-of-the-art (SOTA) methods address this limitation through specialized filtering strategies. For instance, U2PL incorporates unreliable pseudo-labels in contrastive training \cite{wang2022semi_unreliable}, while ELN employs an error-localization network to identify invalid masks \cite{han2024deep}. UniMatch streamlines the process by utilizing a single-stage pipeline with multiple robust augmentation branches \cite{yang2023unimatch}. However, most current frameworks continue to rely on local consistency or pixel-independent thresholding, thereby overlooking complex semantic relationships within unlabeled data \cite{sun2023corrmatchlabelpropagationcorrelation, pham2025learning}.

\subsection{Multimodal Learning in Pathology}
Recent developments in computational pathology have shifted from traditional vision-only frameworks toward multimodal foundation models that integrate visual morphology with complex pathological semantics \cite{lu2024visual, chen2024towards, fu2025multimodal}. Early semi-supervised methods, including UAMT \cite{yu2019uncertainty}, FixMatch \cite{sohn2020fixmatch}, and CPS \cite{chen2021cps}, emphasized visual consistency and perturbation-based invariance to stabilize predictions. These methods, however, are limited by the pronounced morphological heterogeneity in histopathology, as they lack explicit semantic anchors to constrain the high-dimensional feature space \cite{maleki2022lile}. As a result, recent architectures increasingly utilize pathology-specific foundation models, such as UNI \cite{chen2024towards}, QuiltNet~\cite{ikezogwo2023quilt1m} and CONCH \cite{lu2024visual}, which are pretrained on large-scale gigapixel datasets to generate robust representations that capture both local cellular features and global tissue composition \cite{campanella2019clinical}. Although models like MI-Zero \cite{lu2023mi_zero} have introduced vision-language alignment for zero-shot tasks, their emphasis on image-level classification limits their ability to provide the fine-grained, pixel-wise guidance needed for dense prediction tasks.

Recent methods such as DuSSS \cite{pan2024dusss}, Text-SemiSeg \cite{huang2025textdrivenmultiplanarvisualinteraction}, and MPAMatch \cite{fu2025multimodal} demonstrate that incorporating class-level semantic guidance can improve pseudo-label refinement and reduce confirmation bias at ambiguous boundaries. Building on this direction, UniSemAlign explores dual-modal semantic alignment to integrate visual and textual information for dense prediction. By aligning text representations from CONCH \cite{lu2024visual} with visual features extracted by a UNI backbone \cite{chen2024towards}, the proposed framework provides complementary semantic cues to better guide learning from unlabeled data.
\begin{figure*}[t]
    \centering
    \includegraphics[
        width=\textwidth,
        trim=0cm 0cm 0cm 0cm,
        clip
    ]{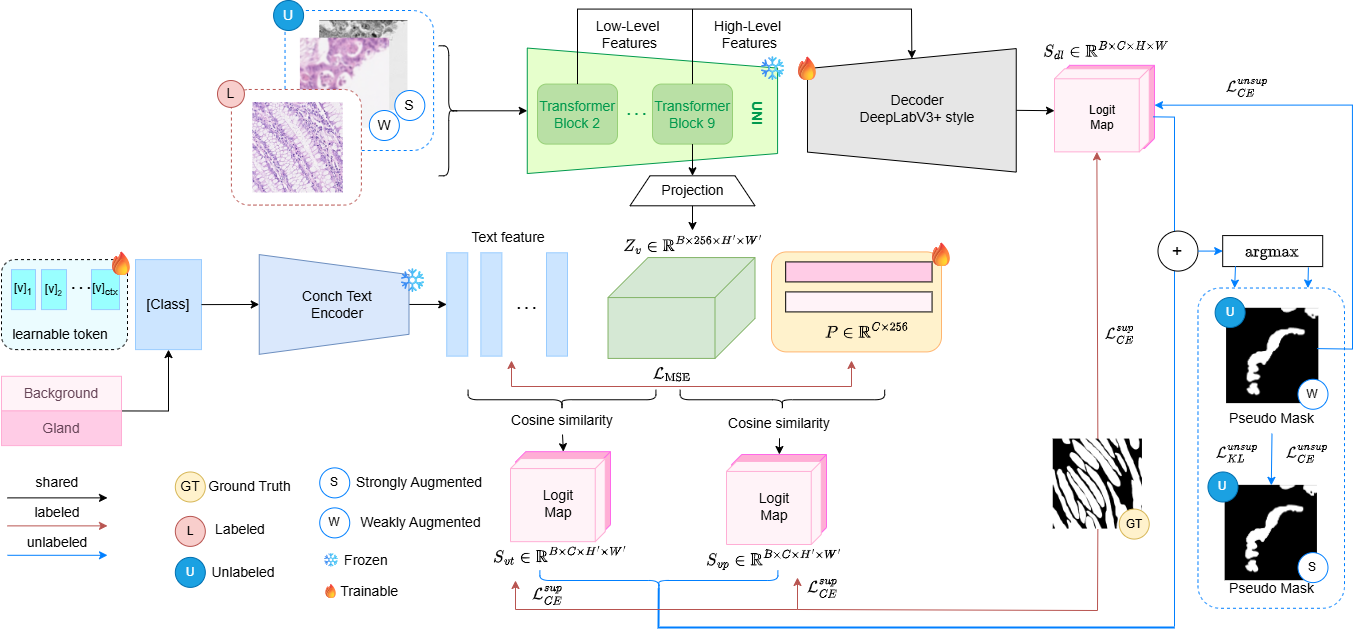}
\caption{Overview of UniSemAlign. 
An input image is encoded by UNI ViT-B/16 and decoded by DeepLabV3+ to produce visual logits. 
In parallel, pixel embeddings are projected into a shared semantic space and aligned with learnable class prototypes and prompt-guided text embeddings generated by a frozen CONCH text encoder with learnable context tokens. 
Cosine similarity between pixel embeddings and these class representations produces alignment logits, which are fused with the decoder logits to generate pseudo-labels for semi-supervised training.}
    \label{fig:framework}
\end{figure*}
\section{Methodology}
\paragraph{Overview.} UniSemAlign performs semi-supervised semantic segmentation for histopathology images by integrating visual prediction with dual-modal semantic alignment. As illustrated in Fig.~\ref{fig:framework}, an input image is encoded by a pathology-pretrained Transformer foundation model (UNI ViT-B/16)~\cite{chen2024towards} to extract hierarchical features, which are passed to a DeepLabV3+~\cite{chen2018encoderdecoderatrousseparableconvolution} decoder to produce visual logits. To inject explicit class-level semantic structure into pixel representations, UniSemAlign introduces two complementary alignment branches operating in a shared embedding space, where high-level pixel features are aligned with learnable class prototypes and text representations derived from class names using a frozen pretrained CONCH ViT-B/16~\cite{lu2024visual} text encoder with learnable context tokens. The resulting dual-modal alignment enforces semantically consistent and discriminative pixel features under limited supervision, and the semantic cues are integrated with visual predictions for both labeled and unlabeled samples during training. The entire framework is optimized end-to-end using supervised segmentation, cross-view consistency, and cross-modal alignment objectives to enhance robustness in data-scarce pathology settings.

\subsection{Visual Segmentation Backbone}

Our segmentation framework follows an encoder--decoder architecture, 
where the encoder is initialized from a pretrained pathology foundation model (UNI ViT-B/16) and the decoder adopts a standard DeepLabV3+ design.

\paragraph{Encoder.}
Given an input image 
\(x \in \mathbb{R}^{B \times 3 \times H \times W}\),
we adopt a pretrained UNI ViT-B/16 Vision Transformer as the image encoder.
The image is partitioned into non-overlapping \(16 \times 16\) patches and processed by Transformer blocks.

We extract intermediate token representations from block 2 and block 9.
After removing the CLS token, the remaining tokens are reshaped into spatial feature maps, denoted as
\(F_2\) and \(F_9\).
These features are projected by \(1 \times 1\) convolutions:
\[
f_{low} = \mathrm{Conv}_{1\times1}(F_2),
\quad
f_{low} \in \mathbb{R}^{B \times 256 \times H' \times W'},
\]
\[
f_{high} = \mathrm{Conv}_{1\times1}(F_9),
\quad
f_{high} \in \mathbb{R}^{B \times 2048 \times H' \times W'},
\]

where \(H' = H/16\) and \(W' = W/16\) correspond to the patch resolution of ViT-B/16.
\paragraph{Decoder.}
We adopt a standard DeepLabV3+ decoder to aggregate multi-scale features. 
The high-level feature $f_{high}$ is processed by an Atrous Spatial Pyramid Pooling (ASPP) module and upsampled, while the low-level feature $f_{low}$ is channel-reduced and fused with the ASPP output. 
The fused feature is mapped to class logits via a $1\times1$ classifier and bilinearly upsampled to the original resolution:
\[
S_{dl} \in \mathbb{R}^{B \times C \times H \times W},
\]
where $C$ denotes the number of segmentation classes.
\subsection{Dual Semantic Alignment Branches}

To inject class-level semantic structure into pixel representations, we introduce two alignment branches that operate in a shared embedding space and produce class logits via cosine similarity: a \emph{prototype branch} and a \emph{text branch}.

\paragraph{Pixel Embedding Projection.}

Given the high-level encoder feature 
\( f_{high} \in \mathbb{R}^{B \times 2048 \times H' \times W'} \),
we apply a $3\times3$ convolutional projection head to map the features into a shared semantic embedding space:
\[
Z = \mathrm{Conv}_{3\times3}(f_{high})
\in \mathbb{R}^{B \times D \times H' \times W'},
\]
where $D=256$.

The embeddings are reshaped and $\ell_2$-normalized along the channel dimension:
\[
z = \mathrm{normalize}(\mathrm{reshape}(Z)),
\quad
z \in \mathbb{R}^{B \times N \times D},
\quad
N = H'W'.
\]

\paragraph{Prototype Branch.}

We maintain learnable class prototypes
\[
P \in \mathbb{R}^{C \times D},
\]
which serve as class-level semantic anchors in the embedding space.
The prototypes are initialized with scaled random weights $(1/\sqrt{D})$.

Prototype logits are computed using cosine similarity between
pixel embeddings and class prototypes:
\[
S_{zp}
=
z\,
\mathrm{normalize}(P)^{\top}
\in
\mathbb{R}^{B \times N \times C}
\]
The logits are reshaped to spatial form:
\[
S_{zp} \in \mathbb{R}^{B \times C \times H' \times W'}.
\]
\paragraph{Text Branch.}

Given $C$ class labels $\{l_1, \dots, l_C\}$, we construct
class-specific prompts following a CoOp-style strategy~\cite{zhou2022learning}.
For each class $c$, we introduce $M$ learnable context tokens
$\{v_1^{(c)}, \dots, v_M^{(c)}\}$ and define the prompt as
\[
\pi_c =
[\text{BOS}]
[v_1^{(c)}]
\dots
[v_M^{(c)}]
[l_c]
[\text{EOT}].
\]
The prompt $\pi_c$ is processed by a pretrained CONCH ViT-B/16
text encoder, which is kept frozen during training.

The token sequence for class $c$ (with placeholder tokens at the context
positions) is first embedded using the CONCH token embedding layer.
The embeddings at the $M$ context positions are then replaced by the
learnable context tokens $\{v_1^{(c)}, \dots, v_M^{(c)}\}$, and positional
embeddings are added to obtain the sequence representation
\[
x_c \in \mathbb{R}^{L \times D_{\text{tok}}},
\]
where $L$ denotes the token sequence length and $D_{\text{tok}}$
is the token embedding dimension of the CONCH text encoder.

The embedded sequence is processed by the frozen CONCH Transformer
text encoder:
\[
h_c =
\mathrm{Encoder}_{\mathrm{CONCH}}(x_c),
\quad
h_c \in
\mathbb{R}^{L \times D_{\text{text}}}.
\]
Following CLIP-style text encoding, we extract the hidden state
corresponding to the end-of-text (EOT) token:
\[
h_{\mathrm{EOT}}^{(c)}
\in
\mathbb{R}^{D_{\text{text}}}.
\]
The EOT representation is first projected by the CONCH text projection
matrix $W_{\mathrm{clip}}$ and then mapped into the shared semantic
embedding space using a learnable linear layer $W_{\mathrm{proj}}$:
\[
f_c^{\text{text}}
=
W_{\mathrm{proj}}
\left(
h_{\mathrm{EOT}}^{(c)}
W_{\mathrm{clip}}
\right),
\quad
f_c^{\text{text}}
\in
\mathbb{R}^{D}.
\]
Stacking all class-wise text features yields the class text embedding
matrix
\[
T =
\left[
f_1^{\text{text}}, \dots, f_C^{\text{text}}
\right]^\top
\in
\mathbb{R}^{C \times D}.
\]
Text alignment logits are computed via cosine similarity:
\[
S_{zt}
=
z\,
\mathrm{normalize}(T)^\top
\in
\mathbb{R}^{B \times N \times C}.
\]

The logits are reshaped to spatial form:
\[
S_{zt}
\in
\mathbb{R}^{B \times C \times H' \times W'}.
\]
\subsection{Logit Fusion for Pseudo-Labeling}
\label{sec:logit_fusion}

Given the decoder logits $S_{dl}$, the semantic alignment branches
produce additional similarity maps.

Since the alignment branches operate at feature resolution
$(H', W')$, their outputs are bilinearly upsampled to the
image resolution:
\[
S_{zp}, S_{zt} \in \mathbb{R}^{B \times C \times H \times W}.
\]

The fused logits for pseudo-label generation are computed as
\[
S_{\mathrm{fuse}}
=
S_{dl}
+
\eta_p S_{zp}
+
\eta_t S_{zt},
\]

where $\eta_p$ and $\eta_t$ are scalar fusion weights controlling the
contributions of the prototype and text alignment branches.
The decoder logits act as the primary prediction, while the semantic
branches provide residual corrections that refine the visual prediction.

\subsection{Semantic-Guided Pseudo-Labeling}

For unlabeled images, we adopt a weak-to-strong consistency framework 
similar to CorrMatch~\cite{sun2023corrmatchlabelpropagationcorrelation}. 
Pseudo-labels are generated from the fused weak-view logits 
$S_{\mathrm{fuse}}^{w}$ (Sec.~\ref{sec:logit_fusion}) via softmax with confidence thresholding. 
The strongly augmented view is then supervised using cross-entropy with hard pseudo-labels (argmax), together with a Kullback--Leibler (KL) divergence term that enforces distribution consistency between weak and strong predictions.

Compared with the baseline framework, our pseudo-labels are derived from semantically guided fused logits that integrate prototype and text alignment, leading to more semantically consistent supervision.
\subsection{Training Objectives}

\paragraph{Overall objective.}
Following CorrMatch~\cite{sun2023corrmatchlabelpropagationcorrelation}, 
we adopt the same loss formulation:
\[
\mathcal{L}
=
\frac{1}{2}
\left(
\mathcal{L}_{\mathrm{sup}}
+
\mathcal{L}_{\mathrm{unsup}}
\right).
\]
where $\mathcal{L}_{\mathrm{sup}}$ and $\mathcal{L}_{\mathrm{unsup}}$ denote 
the supervised and unsupervised objectives, respectively.
\paragraph{Supervised losses.}
For labeled images with ground-truth mask 
\( y \in \{0,\ldots,C-1\}^{B \times H \times W} \) 
(where 255 denotes ignored pixels), 
the supervised objective consists of the standard decoder segmentation loss together with additional supervision on the semantic alignment branches.

The decoder is optimized using pixel-wise cross-entropy:
\[
\mathcal{L}^{\mathrm{sup}}_{\mathrm{dl}} = \mathrm{CE}(S_{dl}, y),
\]
where $S_{dl} \in \mathbb{R}^{B \times C \times H \times W}$ denotes the decoder logits
and $\mathrm{CE}(\cdot)$ is the pixel-wise cross-entropy loss.

Similarly, the prototype-based and text-based logits 
$S_{zp}, S_{zt} \in \mathbb{R}^{B \times C \times H \times W}$ 
are supervised using the same objective:
\[
\mathcal{L}^{\mathrm{sup}}_{\mathrm{proto}} = \mathrm{CE}(S_{zp}, y),
\qquad
\mathcal{L}^{\mathrm{sup}}_{\mathrm{text}} = \mathrm{CE}(S_{zt}, y).
\]

To encourage cross-modal semantic consistency at the class level,
we additionally minimize
\[
\mathcal{L}_{\mathrm{align}}
=
\mathrm{MSE}\bigl(
\mathrm{normalize}(P),
\mathrm{normalize}(T)
\bigr),
\]
where $P, T \in \mathbb{R}^{C \times D}$ denote the prototype and text embeddings,
and $\mathrm{MSE}(\cdot)$ is the mean squared error loss.

The overall supervised objective is defined as
\[
\mathcal{L}_{\mathrm{sup}}
=
\mathcal{L}^{\mathrm{sup}}_{\mathrm{dl}}
+
\mathcal{L}^{\mathrm{sup}}_{\mathrm{proto}}
+
\mathcal{L}^{\mathrm{sup}}_{\mathrm{text}}
+
\mathcal{L}_{\mathrm{align}}.
\]

\paragraph{Unsupervised losses.}

Given a mini-batch of $K$ unlabeled images
$\{u_i\}_{i=1}^{K}$, we generate weak and strong
augmented views, denoted as $u_i^w$ and $u_i^s$.
Pseudo-labels are obtained from the fused logits
$S_{\mathrm{fuse}}^{w}$ (Sec.~\ref{sec:logit_fusion})
via softmax and confidence thresholding,
yielding hard pseudo-labels $\hat{y}_i$.
Low-confidence pixels are ignored during loss computation.

These pseudo-labels serve as supervision targets,
while the unsupervised losses are computed using the decoder
predictions $S_{dl}$, treating the fused prediction as a
stronger teacher signal.

We adopt the same unsupervised learning strategy as CorrMatch,
which consists of three objectives.

\textbf{Hard loss.}
\[
\mathcal{L}^{\mathrm{unsup}}_{h}
=
\frac{1}{K}
\sum_{i=1}^{K}
\mathrm{CE}\big(
S_{dl}(u_i^s), \hat{y}_i
\big).
\]

\textbf{Soft consistency loss.}
\[
\mathcal{L}^{\mathrm{unsup}}_{s}
=
\frac{1}{K}
\sum_{i=1}^{K}
\mathrm{KL}\big(
\sigma(S_{dl}(u_i^s))
\;\|\;
\sigma(S_{dl}(u_i^w))
\big),
\]
where $\sigma(\cdot)$ denotes the softmax operator and
$\mathrm{KL}(\cdot)$ denotes the pixel-wise Kullback--Leibler divergence.

\textbf{Correlation loss.}
\[
\mathcal{L}^{\mathrm{unsup}}_{c}
=
\frac{1}{K}
\sum_{i=1}^{K}
\mathrm{CE}\big(
S_{dl}^{\mathrm{fp}}(u_i^w),
\hat{y}_i
\big),
\]
where $S_{dl}^{\mathrm{fp}}$ denotes the prediction under feature perturbation.

The overall unsupervised objective is defined as:
\[
\mathcal{L}_{\mathrm{unsup}}
=
\lambda_h \mathcal{L}^{\mathrm{unsup}}_{h}
+
\lambda_s \mathcal{L}^{\mathrm{unsup}}_{s}
+
\lambda_c \mathcal{L}^{\mathrm{unsup}}_{c}.
\]

We use the same loss weights as CorrMatch:
$[\lambda_h, \lambda_s, \lambda_c] = [0.5, 0.25, 0.25]$.

\section{Experiments}
\subsection{Experiment Setup}
\paragraph{Datasets.}
The proposed method is evaluated on two benchmark histopathological gland segmentation datasets: \textbf{CRAG}~\cite{graham2019mild} and \textbf{GlaS}~\cite{sirinukunwattana2017gland}. 
CRAG contains 213 high-resolution colon histopathology images (typically $1512 \times 1516$ pixels), where gland segmentation is particularly challenging due to substantial variations in gland morphology and staining patterns. 
GlaS comprises 165 colorectal adenocarcinoma images collected from different cancer stages, following the standard protocol with 80 images reserved for testing. 
All experiments are conducted under semi-supervised settings with 10\% and 20\% labeled data, while the remaining training images are treated as unlabeled samples.

\begin{table*}[!t]
\centering
\caption{Quantitative results on GlaS-2017 and CRAG-2019 datasets under two labeled
ratios. We highlight the best performance for each metric in \textbf{bold}, and the second-best in \underline{underlined}.}
\label{tab:results}
\small
\begin{tabular}{llcccc}
\toprule
\multirow{2}{*}{\textbf{Labeled Ratio}} & \multirow{2}{*}{\textbf{Method}} 
& \multicolumn{2}{c}{\textbf{GlaS}} 
& \multicolumn{2}{c}{\textbf{CRAG}} \\ 
\cmidrule(lr){3-4} \cmidrule(lr){5-6}
& & \textbf{mDice (\%) $\uparrow$} & \textbf{mJaccard (\%) $\uparrow$} 
& \textbf{mDice (\%) $\uparrow$} & \textbf{mJaccard (\%) $\uparrow$}  \\ 
\midrule

100\% 
& Fully-Supervised 
& 89.97 & 81.77 
& 89.13 & 80.44 \\ 
\midrule

\multirow{9}{*}{10\%}
& UAMT (MICCAI 2019) \cite{yu2019uncertainty} & 78.57 & 64.70 & 77.85 & 63.74 \\
& FixMatch (NeurIPS 2020) \cite{sohn2020fixmatch} & 66.01 & 51.39 & 72.42 & 64.31 \\
& CPS (CVPR 2021) \cite{ouali2020semi} & 61.70 & 49.34 & 47.10 & 40.77 \\
& CT (MIDL 2022) \cite{luo2021semi} & 81.75 & 70.44 & 77.33 & 65.56\\
& XNet (ICCV 2023) \cite{xnetv2_2023} & 74.13 & 58.90 & 65.27 & 50.40 \\
& CorrMatch (CVPR 2024) \cite{sun2023corrmatchlabelpropagationcorrelation} & \underline{85.54} & \underline{74.74} & \underline{79.93} & 66.76 \\
& DuSSS (AAAI 2025) \cite{pan2024dusss} & 75.07 & 61.46 & 64.25 & 49.99 \\
& CSDS (MICCAI 2025) \cite{pham2025learning} & 82.89 & 71.61 & 79.86 & \underline{67.81} \\ 
\cmidrule{2-6}
& \textbf{UniSemAlign (Ours)} 
& \textbf{88.15} & \textbf{78.82} 
& \textbf{88.57} & \textbf{79.52} \\ 
\midrule

\multirow{9}{*}{20\%}
& UAMT (MICCAI 2019) \cite{yu2019uncertainty} & 77.09 & 62.72 & 79.76 & 66.33 \\
& FixMatch (NeurIPS 2020) \cite{sohn2020fixmatch} & 58.51 & 44.12 & 71.20 & 64.75 \\
& CPS (CVPR 2021) \cite{ouali2020semi} & 75.99 & 63.69 & 52.43 & 45.70 \\
& CT (MIDL 2022) \cite{luo2021semi} & 85.96 & \underline{76.51} & \underline{81.64} & \underline{70.87} \\
& XNet (ICCV 2023) \cite{xnetv2_2023} & 79.56 & 67.50 & 65.60 & 50.85 \\
& CorrMatch (CVPR 2024) \cite{sun2023corrmatchlabelpropagationcorrelation} & \underline{86.09} & 75.59 & 85.35 & 74.54 \\
& DuSSS (AAAI 2025) \cite{pan2024dusss} & 79.50 & 66.80 & 71.13 & 57.45 \\
& CSDS (MICCAI 2025) \cite{pham2025learning} & 83.50 & 72.87 & 81.28 & 69.67 \\ 
\cmidrule{2-6}
& \textbf{UniSemAlign (Ours)} 
& \textbf{88.58} & \textbf{79.50} 
& \textbf{89.40} & \textbf{80.88} \\ 
\bottomrule
\end{tabular}
\end{table*}

\paragraph{Implementation Details.}
We implemented our method in PyTorch and conducted all experiments on a single NVIDIA TITAN Xp GPU (12GB memory). 
The encoder is a frozen pre-trained UNI ViT-B/16, and the decoder follows DeepLabV3+. 
We use four learnable context tokens in the text encoder and set the fusion weights $\eta_p$ and $\eta_t$ to 0.1. 
Training is performed for 80 epochs using SGD (momentum 0.9, weight decay $10^{-4}$) with an initial learning rate of 0.001 and polynomial decay. 
Input images are randomly rescaled and cropped to $256 \times 256$. 
For unlabeled data, we follow CorrMatch~\cite{sun2023corrmatchlabelpropagationcorrelation} for weak--strong augmentation and feature perturbation, and additionally apply CutMix~\cite{yun2019cutmixregularizationstrategytrain}. 
Pseudo-labels are generated from fused logits using an EMA-updated confidence threshold initialized at 0.7. 
Evaluation is conducted with single-scale or sliding-window inference depending on the dataset, and performance is measured by Dice and Jaccard metrics. 

\subsection{Quantitative Results}

We compare UniSemAlign with a diverse set of semi-supervised segmentation baselines across multiple SSL paradigms, including generic SSL methods (e.g., FixMatch \cite{sohn2020fixmatch}, CPS \cite{ouali2020semi}, CT \cite{luo2021semi}), correlation-based refinement (CorrMatch \cite{sun2023corrmatchlabelpropagationcorrelation}), and medical SSL approaches (UAMT \cite{yu2019uncertainty}). We also include recent pathology-related methods such as CSDS \cite{pham2025learning} and XNet v2 \cite{xnetv2_2023}, as well as DuSSS \cite{pan2024dusss}, a recent semi-supervised method that incorporates textual information for semantic guidance. The quantitative results are summarized in Table~\ref{tab:results}.

On \textbf{GlaS-2017}, UniSemAlign outperforms the strongest baseline CorrMatch by $+2.61\%$ mDice and $+4.08\%$ mJaccard under 10\% labeling, and by $+2.49\%$ mDice and $+2.99\%$ mJaccard under 20\% labeling. On the more challenging \textbf{CRAG-2019} dataset, the improvements are substantially larger, reaching $+8.64\%$ mDice and $+11.71\%$ mJaccard at 10\% labeling, and $+7.76\%$ mDice and $+10.01\%$ mJaccard at 20\% labeling over the strongest prior results. These gains indicate that incorporating prototype-based and text-based semantic constraints improves pseudo-label refinement, particularly in complex pathology structures.

\subsection{Qualitative Results}

\begin{figure*}[t]
    \centering
    \includegraphics[
        width=\textwidth,
        trim=0cm 0cm 0cm 0cm,
        clip
    ]{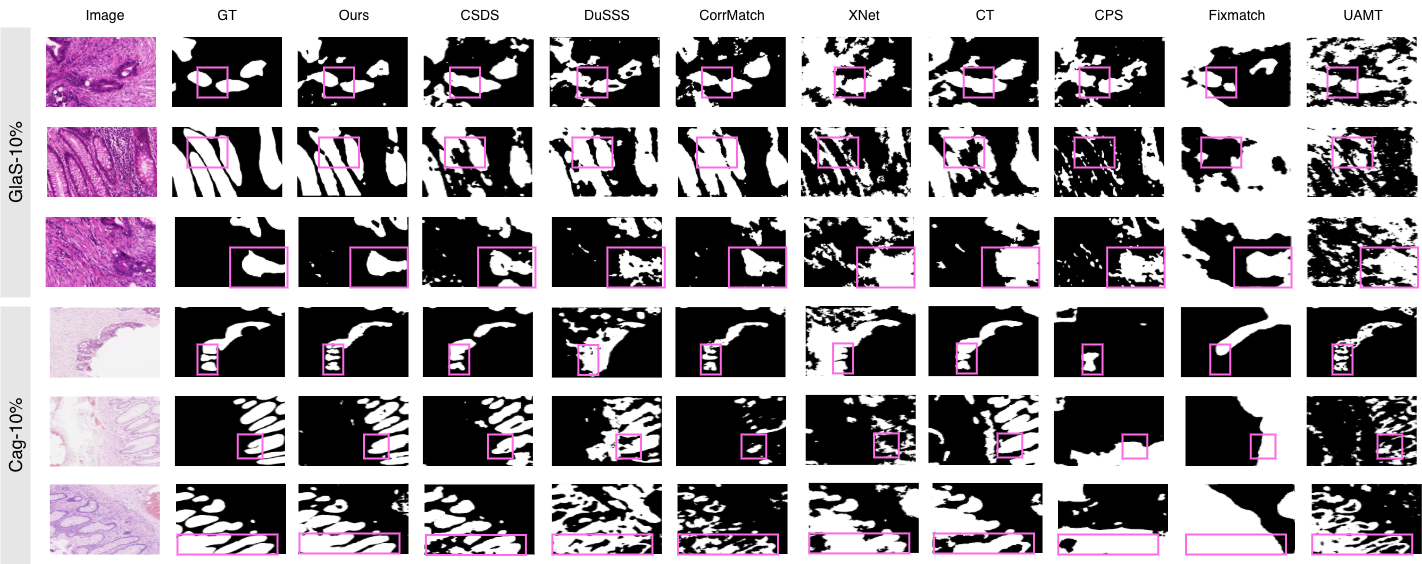}
    \caption{Qualitative results for different semi-supervised methods under the 10\% labeling setting on GlaS-2017 and CRAG-2019. Pink boxes highlight notable differences between competing methods and our approach.}
    \label{fig:qualitative}
\end{figure*}

Fig.~\ref{fig:qualitative} presents qualitative comparisons under the 10\% labeling setting on GlaS-2017 and CRAG-2019. On GlaS, many competing methods produce noisy predictions with broken boundaries or missing gland parts, especially in low-contrast regions, leading to fragmented structures and incomplete gland shapes. Several baselines also introduce spurious foreground regions or holes inside the predicted glands. 
On the more challenging CRAG dataset, baseline methods often suffer from severe over-segmentation or incorrectly merge adjacent glands, producing scattered predictions and large false-positive regions. In contrast, UniSemAlign generates cleaner and more coherent segmentations with smoother boundaries. As highlighted in the boxed regions, our method better preserves thin gland structures, maintains elongated gland shapes, and effectively suppresses isolated noise, resulting in predictions that are closer to the ground truth.

\subsection{Ablation Studies}
\paragraph{Effectiveness of Encoder Architecture.}
As shown in Table~\ref{tab:ablation_encoder}, replacing the ResNet101~\cite{he2015deepresiduallearningimage} encoder with the pathology-pretrained UNI encoder in DeepLabV3+ yields consistent improvements of +1.77 mDice and +2.79 mJaccard. These gains highlight the benefit of domain-specific foundation pretraining for more robust feature representations under limited supervision.
\begin{table}[H]
\centering
\caption{Ablation study on encoder architectures of DeepLabV3+ on GlaS (10\% labeled). The best results are shown in \textbf{bold}.}
\label{tab:ablation_encoder}
\small
\begin{tabular}{lcc}
\toprule
\textbf{Encoder} & \textbf{mDice (\%)} & \textbf{mJaccard (\%)} \\
\midrule
ResNet101 & 86.37 & 76.01 \\
UNI & \textbf{88.15} & \textbf{78.82} \\
\bottomrule
\end{tabular}
\end{table}
\paragraph{Impact of Context Token Number.}
Adopting learnable context tokens improves performance over the no-context setting, with gains of up to +0.62 mDice and +1.00 mJaccard, as shown in Table~\ref{tab:ablation_ctx}. 
Increasing the token number from 2 to 4 yields steady improvements, while additional tokens result in a slight drop. 
Hence, we use 4 context tokens in all experiments.
\begin{table}[t]
\centering
\caption{Impact of the number of learnable context tokens on GlaS (10\% labeled). The best results are shown in \textbf{bold}.}
\label{tab:ablation_ctx}
\small
\begin{tabular}{ccc}
\toprule
\textbf{\# Context Tokens} & \textbf{mDice (\%)} & \textbf{mJaccard (\%)} \\
\midrule
0 & 87.52 & 77.80 \\
2 & 87.85 & 77.86 \\
3 & 88.07 & 78.68 \\
4 & \textbf{88.15} & \textbf{78.82} \\
5 & 87.79 & 78.25 \\
\bottomrule
\end{tabular}
\end{table}

\begin{figure}[H]
    \centering
    \includegraphics[
        width=\linewidth,
        trim=0cm 0cm 0cm 0cm,
        clip
    ]{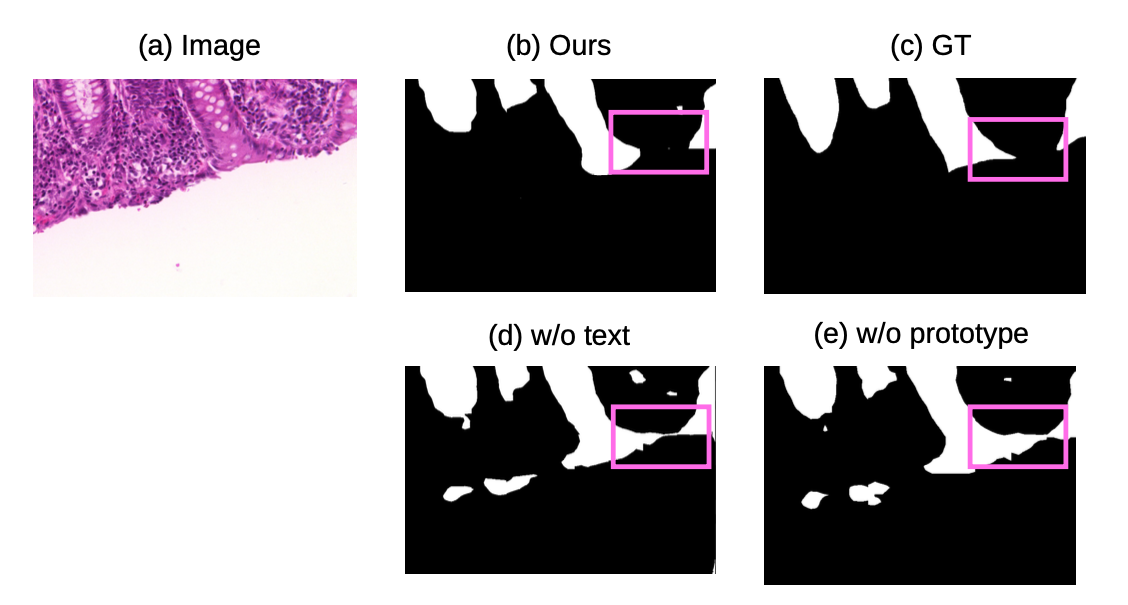}
\caption{Qualitative comparison of the dual semantic alignment branches on GlaS (10\% labeled). From left to right: (a) input image, (b) UniSemAlign (ours), (c) Ground truth, (d) without text branch, (e) without prototype branch. Pink boxes highlight notable differences. The prototype branch primarily improves structural completeness and boundary coherence, while the text branch suppresses ambiguous/background activations. }
    \label{fig:ablation_mask}
\end{figure}

\paragraph{Dual Semantic Alignment Branches Analysis.}

We evaluate the contribution of the text-driven and prototype-based alignment branches (Table~\ref{tab:ablation_glas}). 
Both branches individually improve over the baseline: prototype alignment brings gains of +0.34 mDice and +0.53 mJaccard, while text guidance yields +0.40 and +0.62, respectively. 
Combining both achieves the best performance with overall improvements of +0.48 mDice and +0.76 mJaccard, suggesting complementary semantic roles.

Qualitative comparisons further illustrate this effect (Fig.~\ref{fig:ablation_mask}). 
Without prototype alignment, the predicted glands exhibit internal discontinuities and small missing regions. 
Removing text guidance introduces spurious small segments and less regular boundaries. 
In contrast, the full model produces denser gland regions with smoother contours, more closely matching the ground truth.

\begin{table}[H]
\centering
\caption{Ablation study of semantic alignment branches on GlaS (10\% labeled).  The best results are shown in \textbf{bold}.}
\label{tab:ablation_glas}
\small
\begin{tabular}{c c c c}
\toprule
\textbf{Text} & \textbf{Prototype} & \textbf{mDice (\%)} & \textbf{mJaccard (\%)} \\
\midrule
$\times$ & $\times$ & 87.66 & 78.04 \\
$\times$ & $\checkmark$ & 88.00 & 78.57 \\
$\checkmark$ & $\times$ & 88.06 & 78.66 \\
$\checkmark$ & $\checkmark$ & \textbf{88.15} & \textbf{78.82} \\
\bottomrule
\end{tabular}
\end{table}

\paragraph{Comparison of Alignment Losses.}
We compare different alignment objectives, including cosine similarity, KL divergence, and MSE (Table~\ref{tab:ablation_align_loss}). 
Among them, MSE achieves the best performance, improving over the no-alignment baseline by +0.58 mDice and +0.93 mJaccard. 
Cosine loss provides marginal gains, while KL divergence slightly degrades performance. 
These results suggest that explicit feature-level regression better enforces semantic consistency between prototypes and text embeddings.

\begin{table}[H]
\centering
\caption{Comparison of alignment loss formulations on GlaS (10\% labeled). The best results are shown in \textbf{bold}.}
\label{tab:ablation_align_loss}
\small
\begin{tabular}{lcc}
\toprule
\textbf{Alignment loss} & \textbf{mDice (\%)} & \textbf{mJaccard (\%)} \\
\midrule
No alignment & 87.56 & 77.87 \\
Cosine similarity & 87.69 & 78.08 \\
KL divergence & 87.43 & 77.67 \\
MSE & \textbf{88.15} & \textbf{78.82} \\
\bottomrule
\end{tabular}
\end{table}


\section{Conclusion}

This paper introduced UniSemAlign, a dual-modal semantic alignment framework for semi-supervised histopathology segmentation. 
By jointly leveraging prototype-based structural alignment and text-driven semantic guidance in a shared embedding space, our method imposes explicit class-level constraints to enhance pixel-wise learning and pseudo-label refinement. 
Built upon a pathology-pretrained foundation encoder, UniSemAlign consistently improves performance under limited annotation settings. Experiments on GlaS-2017 and CRAG-2019 demonstrate not only quantitative gains but also more structurally coherent gland segmentation. Overall, our results suggest that integrating visual prototypes with textual semantics provides an effective and scalable strategy for semi-supervised medical image segmentation.

\section{Acknowledgement}

We thank AI VIETNAM for facilitating computational resources and financial support for this paper.

{
    \small
    \bibliographystyle{ieeenat_fullname}
    \bibliography{main}

@String(CVPR= {IEEE Conf. Comput. Vis. Pattern Recog.})

@String(BMVC= {Brit. Mach. Vis. Conf.})

@String(CVPR  = {CVPR})

@String(BMVC  =	{BMVC})

@article{pham2025learning,
  title={Learning disentangled stain and structural representations for semi-supervised histopathology segmentation},
  author={Pham, Ha-Hieu and Vu, Nguyen Lan Vi and Nguyen, Thanh-Huy and Bagci, Ulas and Xu, Min and Le, Trung-Nghia and Pham, Huy-Hieu},
  journal={arXiv preprint arXiv:2507.03923},
  year={2025}
}

@article{sirinukunwattana2017gland,
  author    = {Sirinukunwattana, K. and Pluim, J. P. W. and Chen, H. and Qi, X. and Heng, P. A. and Guo, Y. B. and Wang, L. Y. and Matuszewski, B. J. and Bruni, E. and Sanchez, U. and others},
  title     = {Gland segmentation in colon histology images: The {GlaS} challenge contest},
  journal   = {Medical Image Analysis},
  volume    = {35},
  pages     = {489--502},
  year      = {2017},
  publisher={Elsevier}
}

@article{pan2024dusss,
  title   = {DuSSS: Dual Semantic Similarity-Supervised Vision-Language Model for Semi-Supervised Medical Image Segmentation},
  author  = {Pan, Qingtao and Qiao, Wenhao and Lou, Jingjiao and Ji, Bing and Li, Shuo},
  journal = {arXiv preprint arXiv:2412.12492},
  year    = {2024},
  doi     = {10.48550/arXiv.2412.12492},
  url     = {https://arxiv.org/abs/2412.12492}
}

@article{graham2019mild,
  author    = {Graham, S. and Chen, H. and Gamper, J. and Dou, Q. and Heng, P. A. and Snead, D. and Tsang, Y. W. and Rajpoot, N.},
  title     = {Mild-net: Minimal information loss dilated network for gland instance segmentation in colon histology images},
  journal   = {Medical Image Analysis},
  volume    = {52},
  pages     = {199--211},
  year      = {2019},
  publisher={Elsevier}
}

@misc{sun2023corrmatchlabelpropagationcorrelation,
      title={CorrMatch: Label Propagation via Correlation Matching for Semi-Supervised Semantic Segmentation}, 
      author={Boyuan Sun and Yuqi Yang and Le Zhang and Ming-Ming Cheng and Qibin Hou},
      year={2023},
      eprint={2306.04300},
      archivePrefix={arXiv},
      primaryClass={cs.CV},
      url={https://arxiv.org/abs/2306.04300}, 
}

@misc{chen2018encoderdecoderatrousseparableconvolution,
      title={Encoder-Decoder with Atrous Separable Convolution for Semantic Image Segmentation}, 
      author={Liang-Chieh Chen and Yukun Zhu and George Papandreou and Florian Schroff and Hartwig Adam},
      year={2018},
      eprint={1802.02611},
      archivePrefix={arXiv},
      primaryClass={cs.CV},
      url={https://arxiv.org/abs/1802.02611}, 
}

@article{frank2023accurate,
  title={Accurate diagnostic tissue segmentation and concurrent disease subtyping with small datasets},
  author={Frank, Steven J},
  journal={Journal of Pathology Informatics},
  volume={14},
  pages={100174},
  year={2023},
  publisher={Elsevier}
}

@inproceedings{hashimoto2020multi,
  title={Multi-scale domain-adversarial multiple-instance CNN for cancer subtype classification with unannotated histopathological images},
  author={Hashimoto, Noriaki and Fukushima, Daisuke and Koga, Ryoichi and Takagi, Yusuke and Ko, Kaho and Kohno, Kei and Nakaguro, Masato and Nakamura, Shigeo and Hontani, Hidekata and Takeuchi, Ichiro},
  booktitle={Proceedings of the IEEE/CVF conference on computer vision and pattern recognition},
  pages={3852--3861},
  year={2020}
}

@inproceedings{li2023scribblevc,
  title={ScribbleVC: Scribble-supervised medical image segmentation with vision-class embedding},
  author={Li, Zihan and Zheng, Yuan and Luo, Xiangde and Shan, Dandan and Hong, Qingqi},
  booktitle={Proceedings of the 31st ACM International Conference on Multimedia},
  pages={3384--3393},
  year={2023}
}

@article{luo2021semi,
  title   = {Semi-Supervised Medical Image Segmentation via Cross Teaching between CNN and Transformer},
  author  = {Luo, Xiangde and Hu, Minhao and Song, Tao and Wang, Guotai and Zhang, Shaoting},
  journal = {arXiv preprint arXiv:2112.04894},
  year    = {2021},
  doi     = {10.48550/arXiv.2112.04894},
  url     = {https://arxiv.org/abs/2112.04894}
}

@inproceedings{zhang2022pixelseg,
  title={Pixelseg: Pixel-by-pixel stochastic semantic segmentation for ambiguous medical images},
  author={Zhang, Wei and Zhang, Xiaohong and Huang, Sheng and Lu, Yuting and Wang, Kun},
  booktitle={Proceedings of the 30th ACM International Conference on Multimedia},
  pages={4742--4750},
  year={2022}
}

@inproceedings{long2015fully,
  title={Fully convolutional networks for semantic segmentation},
  author={Long, Jonathan and Shelhamer, Evan and Darrell, Trevor},
  booktitle={Proceedings of the IEEE conference on computer vision and pattern recognition},
  pages={3431--3440},
  year={2015}
}

@inproceedings{ronneberger2015u,
  title={U-net: Convolutional networks for biomedical image segmentation},
  author={Ronneberger, Olaf and Fischer, Philipp and Brox, Thomas},
  booktitle={International Conference on Medical image computing and computer-assisted intervention},
  pages={234--241},
  year={2015},
  organization={Springer}
}

@article{chen2021transunet,
  title={Transunet: Transformers make strong encoders for medical image segmentation},
  author={Chen, Jieneng and Lu, Yongyi and Yu, Qihang and Luo, Xiangde and Adeli, Ehsan and Wang, Yan and Lu, Le and Yuille, Alan L and Zhou, Yuyin},
  journal={arXiv preprint arXiv:2102.04306},
  year={2021}
}

@inproceedings{cao2022swin,
  title={Swin-unet: Unet-like pure transformer for medical image segmentation},
  author={Cao, Hu and Wang, Yueyue and Chen, Joy and Jiang, Dongsheng and Zhang, Xiaopeng and Tian, Qi and Wang, Manning},
  booktitle={European conference on computer vision},
  pages={205--218},
  year={2022},
  organization={Springer}
}

@article{campanella2019clinical,
  title={Clinical-grade computational pathology using weakly supervised deep learning on whole slide images},
  author={Campanella, Gabriele and Hanna, Matthew G and Geneslaw, Luke and Miraflor, Allen and Werneck Krauss Silva, Vitor and Busam, Klaus J and Brogi, Edi and Reuter, Victor E and Klimstra, David S and Fuchs, Thomas J},
  journal={Nature medicine},
  volume={25},
  number={8},
  pages={1301--1309},
  year={2019},
  publisher={Nature Publishing Group US New York}
}

@article{zhang2024dslsm,
  title={DSLSM: Dual-kernel-induced statistic level set model for image segmentation},
  author={Zhang, Fan and Liu, Huiying and Duan, Xiaojun and Wang, Binglu and Cai, Qing and Li, Huafeng and Dong, Junyu and Zhang, David},
  journal={Expert Systems with Applications},
  volume={242},
  pages={122772},
  year={2024},
  publisher={Elsevier}
}

@article{han2024deep,
  title={Deep semi-supervised learning for medical image segmentation: A review},
  author={Han, Kai and Sheng, Victor S and Song, Yuqing and Liu, Yi and Qiu, Chengjian and Ma, Siqi and Liu, Zhe},
  journal={Expert Systems with Applications},
  volume={245},
  pages={123052},
  year={2024},
  publisher={Elsevier}
}

@article{tarvainen2017mean,
  title={Mean teachers are better role models: Weight-averaged consistency targets improve semi-supervised deep learning results},
  author={Tarvainen, Antti and Valpola, Harri},
  journal={Advances in neural information processing systems},
  volume={30},
  year={2017}
}

@inproceedings{yu2019uncertainty,
  title={Uncertainty-aware self-ensembling model for semi-supervised 3D left atrium segmentation},
  author={Yu, Lequan and Wang, Shujun and Li, Xiaomeng and Fu, Chi-Wing and Heng, Pheng-Ann},
  booktitle={International conference on medical image computing and computer-assisted intervention},
  pages={605--613},
  year={2019},
  organization={Springer}
}

@inproceedings{bai2023bidirectional,
  title={Bidirectional copy-paste for semi-supervised medical image segmentation},
  author={Bai, Yunhao and Chen, Duowen and Li, Qingli and Shen, Wei and Wang, Yan},
  booktitle={Proceedings of the IEEE/CVF conference on computer vision and pattern recognition},
  pages={11514--11524},
  year={2023}
}

@article{sohn2020fixmatch,
  title={Fixmatch: Simplifying semi-supervised learning with consistency and confidence},
  author={Sohn, Kihyuk and Berthelot, David and Carlini, Nicholas and Zhang, Zizhao and Zhang, Han and Raffel, Colin A and Cubuk, Ekin Dogus and Kurakin, Alexey and Li, Chun-Liang},
  journal={Advances in neural information processing systems},
  volume={33},
  pages={596--608},
  year={2020}
}

@inproceedings{xie2022clims,
  title={Clims: Cross language image matching for weakly supervised semantic segmentation},
  author={Xie, Jinheng and Hou, Xianxu and Ye, Kai and Shen, Linlin},
  booktitle={Proceedings of the IEEE/CVF conference on computer vision and pattern recognition},
  pages={4483--4492},
  year={2022}
}

@inproceedings{radford2021learning,
  title={Learning transferable visual models from natural language supervision},
  author={Radford, Alec and Kim, Jong Wook and Hallacy, Chris and Ramesh, Aditya and Goh, Gabriel and Agarwal, Sandhini and Sastry, Girish and Askell, Amanda and Mishkin, Pamela and Clark, Jack and others},
  booktitle={International conference on machine learning},
  pages={8748--8763},
  year={2021},
  organization={PmLR}
}

@article{huang2023visual,
  title={A visual--language foundation model for pathology image analysis using medical twitter},
  author={Huang, Zhi and Bianchi, Federico and Yuksekgonul, Mert and Montine, Thomas J and Zou, James},
  journal={Nature medicine},
  volume={29},
  number={9},
  pages={2307--2316},
  year={2023},
  publisher={Nature Publishing Group US New York}
}

@article{lu2024visual,
  title={A visual-language foundation model for computational pathology},
  author={Lu, Ming Y and Chen, Bowen and Williamson, Drew FK and Chen, Richard J and Liang, Ivy and Ding, Tong and Jaume, Guillaume and Odintsov, Igor and Le, Long Phi and Gerber, Georg and others},
  journal={Nature medicine},
  volume={30},
  number={3},
  pages={863--874},
  year={2024},
  publisher={Nature Publishing Group US New York}
}

@article{zhou2022learning,
  title={Learning to prompt for vision-language models},
  author={Zhou, Kaiyang and Yang, Jingkang and Loy, Chen Change and Liu, Ziwei},
  journal={International journal of computer vision},
  volume={130},
  number={9},
  pages={2337--2348},
  year={2022},
  publisher={Springer}
}

@article{zhu2005semi,
  title={Semi-supervised learning literature survey},
  author={Zhu, Xiaojin Jerry},
  year={2005},
  publisher={University of Wisconsin-Madison Department of Computer Sciences}
}

@article{ouali2020semi,
  title={Semi-supervised semantic segmentation with cross-consistency training},
  author={Ouali, Yassine and Hudelot, C{\'e}line and Tami, Myriam},
  journal={Proceedings of the IEEE/CVF Conference on Computer Vision and Pattern Recognition (CVPR)},
  pages={12674--12684},
  year={2020}
}

@article{french2020semi,
  title={Semi-supervised semantic segmentation needs strong, varied perturbations},
  author={French, Geoffrey and Laine, Samuli and Aila, Timo and Mackiewicz, Michal and Finlayson, Graham},
  journal={British Machine Vision Conference (BMVC)},
  year={2020}
}

@article{wang2022semi_unreliable,
  title={Semi-supervised semantic segmentation using unreliable pseudo-labels},
  author={Wang, Yuchao and Wang, Haochen and Shen, Yujun and Fei, Jingjing and Li, Wei and Jin, Guoqiang and Wu, Liwei and Zhao, Rui and Le, Xinyi},
  journal={Proceedings of the IEEE/CVF Conference on Computer Vision and Pattern Recognition (CVPR)},
  pages={4248--4257},
  year={2022}
}

@article{vu2025semi,
  title={Semi-MoE: Mixture-of-experts meets semi-supervised histopathology segmentation},
  author={Vu, Nguyen Lan Vi and Nguyen, Thanh-Huy and Nguyen, Thien and Kihara, Daisuke and Wang, Tianyang and Li, Xingjian and Xu, Min},
  journal={arXiv preprint arXiv:2509.13834},
  year={2025}
}

@article{chen2024towards,
  title={Towards a general-purpose foundation model for computational pathology},
  author={Chen, Richard J and Ding, Tong and Lu, Ming Y and Williamson, Drew FK and Jaume, Guillaume and Song, Andrew H and Chen, Bowen and Zhang, Andrew and Shao, Daniel and Shaban, Muhammad and others},
  journal={Nature medicine},
  volume={30},
  number={3},
  pages={850--862},
  year={2024},
  publisher={Nature Publishing Group US New York}
}

@article{fu2025multimodal,
  title={Multimodal Prototype Alignment for Semi-supervised Pathology Image Segmentation},
  author={Fu, Mingxi and Fu, Fanglei and Ling, Xitong and Yuan, Huaitian and Guan, Tian and He, Yonghong and Zhu, Lianghui},
  journal={arXiv preprint arXiv:2508.19574},
  year={2025}
}

@article{wang2021annotation,
  title={Annotation-efficient deep learning for automatic medical image segmentation},
  author={Wang, Shanshan and Li, Cheng and Wang, Rongpin and Liu, Zaiyi and Wang, Meiyun and Tan, Hongna and Wu, Yaping and Liu, Xinfeng and Sun, Hui and Yang, Rui and others},
  journal={Nature communications},
  volume={12},
  number={1},
  pages={5915},
  year={2021},
  publisher={Nature Publishing Group UK London}
}

@article{wang2022multi,
  title={Multi-granularity cross-modal alignment for generalized medical visual representation learning},
  author={Wang, Fuying and Zhou, Yuyin and Wang, Shujun and Vardhanabhuti, Varut and Yu, Lequan},
  journal={Advances in neural information processing systems},
  volume={35},
  pages={33536--33549},
  year={2022}
}

@inproceedings{zhou2024knowledge,
  title={Knowledge-enhanced visual-language pretraining for computational pathology},
  author={Zhou, Xiao and Zhang, Xiaoman and Wu, Chaoyi and Zhang, Ya and Xie, Weidi and Wang, Yanfeng},
  booktitle={European Conference on Computer Vision},
  pages={345--362},
  year={2024},
  organization={Springer}
}

@misc{yun2019cutmixregularizationstrategytrain,
      title={CutMix: Regularization Strategy to Train Strong Classifiers with Localizable Features}, 
      author={Sangdoo Yun and Dongyoon Han and Seong Joon Oh and Sanghyuk Chun and Junsuk Choe and Youngjoon Yoo},
      year={2019},
      eprint={1905.04899},
      archivePrefix={arXiv},
      primaryClass={cs.CV},
      url={https://arxiv.org/abs/1905.04899}, 
}

@inproceedings{maleki2022lile,
  title={LILE: Look in-depth before looking elsewhere--a dual attention network using transformers for cross-modal information retrieval in histopathology archives},
  author={Maleki, David and Tizhoosh, H.R.},
  booktitle={International Conference on Medical Imaging with Deep Learning (MIDL)},
  pages={879--894},
  year={2022},
  organization={PMLR}
}

@inproceedings{lu2023mi_zero,
  title={Visual Language Pretrained Multiple Instance Zero-Shot Transfer for Histopathology Images},
  author={Lu, Ming Y and Chen, Bowen and Zhang, Andrew and Williamson, Drew FK and Chen, Richard J and Ding, Tong and Le, Long Phi and Chuang, Yung-Sung and Mahmood, Faisal},
  booktitle=CVPR,
  pages={19764--19775},
  year={2023}
}

@inproceedings{chen2021cps,
  title={Semi-Supervised Semantic Segmentation with Cross Pseudo Supervision},
  author={Chen, Xiaokang and Yuan, Yuhui and Zeng, Gang and Wang, Jingdong},
  booktitle=CVPR,
  pages={2613--2622},
  year={2021}
}

@inproceedings{yang2023unimatch,
  title={UniMatch: Revisiting Weak-to-Strong Consistency in Semi-Supervised Semantic Segmentation},
  author={Yang, Lihe and Qi, Lei and Feng, Litong and Zhang, Wayne and Shi, Yinghuan},
  booktitle=CVPR,
  pages={7236--7246},
  year={2023}
}

@article{xnetv2_2023,
  title   = {XNet v2: Fewer Limitations, Better Results and Greater Universality},
  author  = {Zhou, Yanfeng and Li, Lingrui and Wang, Zichen and Liu, Guole and Liu, Ziwen and Yang, Ge},
  journal = {arXiv preprint arXiv:2409.00947},
  year    = {2024},
  doi     = {10.48550/arXiv.2409.00947},
  url     = {https://arxiv.org/abs/2409.00947}
}

@misc{he2015deepresiduallearningimage,
      title={Deep Residual Learning for Image Recognition}, 
      author={Kaiming He and Xiangyu Zhang and Shaoqing Ren and Jian Sun},
      year={2015},
      eprint={1512.03385},
      archivePrefix={arXiv},
      primaryClass={cs.CV},
      url={https://arxiv.org/abs/1512.03385}, 
}

@article{shen2023co,
  title={Co-training with high-confidence pseudo labels for semi-supervised medical image segmentation},
  author={Shen, Zhiqiang and Cao, Peng and Yang, Hua and Liu, Xiaoli and Yang, Jinzhu and Zaiane, Osmar R},
  journal={arXiv preprint arXiv:2301.04465},
  year={2023}
}

@article{qin2024urca,
  title={URCA: Uncertainty-based region clipping algorithm for semi-supervised medical image segmentation},
  author={Qin, Chendong and Wang, Yongxiong and Zhang, Jiapeng},
  journal={Computer Methods and Programs in Biomedicine},
  volume={254},
  pages={108278},
  year={2024},
  publisher={Elsevier}
}

@article{nadeem2025segmenting,
  title={Segmenting Visuals With Querying Words: Language Anchors For Semi-Supervised Image Segmentation},
  author={Nadeem, Numair and Anwar, Saeed and Asad, Muhammad Hamza and Bais, Abdul},
  journal={arXiv preprint arXiv:2506.13925},
  year={2025}
}

@inproceedings{kapse2024si,
  title={Si-mil: Taming deep mil for self-interpretability in gigapixel histopathology},
  author={Kapse, Saarthak and Pati, Pushpak and Das, Srijan and Zhang, Jingwei and Chen, Chao and Vakalopoulou, Maria and Saltz, Joel and Samaras, Dimitris and Gupta, Rajarsi R and Prasanna, Prateek},
  booktitle={Proceedings of the IEEE/CVF Conference on Computer Vision and Pattern Recognition},
  pages={11226--11237},
  year={2024}
}

@inproceedings{wu2025learning,
  title={Learning heterogeneous tissues with mixture of experts for gigapixel whole slide images},
  author={Wu, Junxian and Chen, Minheng and Ke, Xinyi and Xun, Tianwang and Jiang, Xiaoming and Zhou, Hongyu and Shao, Lizhi and Kong, Youyong},
  booktitle={Proceedings of the Computer Vision and Pattern Recognition Conference},
  pages={5144--5153},
  year={2025}
}

@article{li2024boundary,
  title={Boundary-aware uncertainty suppression for semi-supervised medical image segmentation},
  author={Li, Congcong and Zhang, Jinshuo and Niu, Dongmei and Zhao, Xiuyang and Yang, Bo and Zhang, Caiming},
  journal={IEEE Transactions on Artificial Intelligence},
  volume={5},
  number={8},
  pages={4074--4086},
  year={2024},
  publisher={IEEE}
}

@inproceedings{xu2024semi,
  title={Semi-supervised segmentation of histopathology images with noise-aware topological consistency},
  author={Xu, Meilong and Hu, Xiaoling and Gupta, Saumya and Abousamra, Shahira and Chen, Chao},
  booktitle={European Conference on Computer Vision},
  pages={271--289},
  year={2024},
  organization={Springer}
}

@article{wang2022ssa,
  title={SSA-Net: Spatial self-attention network for COVID-19 pneumonia infection segmentation with semi-supervised few-shot learning},
  author={Wang, Xiaoyan and Yuan, Yiwen and Guo, Dongyan and Huang, Xiaojie and Cui, Ying and Xia, Ming and Wang, Zhenhua and Bai, Cong and Chen, Shengyong},
  journal={Medical image analysis},
  volume={79},
  pages={102459},
  year={2022},
  publisher={Elsevier}
}

@inproceedings{thompson2022pseudo,
  title={Pseudo-label refinement using superpixels for semi-supervised brain tumour segmentation},
  author={Thompson, Bethany H and Di Caterina, Gaetano and Voisey, Jeremy P},
  booktitle={2022 IEEE 19th International Symposium on Biomedical Imaging (ISBI)},
  pages={1--5},
  year={2022},
  organization={IEEE}
}

@inproceedings{he2025trustmatch,
  title={TrustMatch: mitigating pseudo-label bias in semi-supervised learning with trust-aware refinement},
  author={He, Hongyang and Hong, Yundi},
  booktitle={Proceedings of the IEEE/CVF International Conference on Computer Vision},
  pages={594--603},
  year={2025}
}

@inproceedings{yi2023simple,
  title={A simple framework for text-supervised semantic segmentation},
  author={Yi, Muyang and Cui, Quan and Wu, Hao and Yang, Cheng and Yoshie, Osamu and Lu, Hongtao},
  booktitle={Proceedings of the IEEE/CVF conference on computer vision and pattern recognition},
  pages={7071--7080},
  year={2023}
}

@inproceedings{wang2022medclip,
  title={Medclip: Contrastive learning from unpaired medical images and text},
  author={Wang, Zifeng and Wu, Zhenbang and Agarwal, Dinesh and Sun, Jimeng},
  booktitle={Proceedings of the 2022 Conference on Empirical Methods in Natural Language Processing},
  pages={3876--3887},
  year={2022}
}

@article{gilal2025pathvlm,
  title={PathVLM-Eval: Evaluation of open vision language models in histopathology},
  author={Gilal, Nauman Ullah and Zegour, Rachida and Al-Thelaya, Khaled and {\"O}zer, Erdener and Agus, Marco and Schneider, Jens and Boughorbel, Sabri},
  journal={Journal of Pathology Informatics},
  volume={18},
  pages={100455},
  year={2025},
  publisher={Elsevier}
}

@article{sikaroudi2023generalization,
  title={Generalization of vision pre-trained models for histopathology},
  author={Sikaroudi, Milad and Hosseini, Maryam and Gonzalez, Ricardo and Rahnamayan, Shahryar and Tizhoosh, HR},
  journal={Scientific reports},
  volume={13},
  number={1},
  pages={6065},
  year={2023},
  publisher={Nature Publishing Group UK London}
}

@misc{ikezogwo2023quilt1m,
      title={Quilt-1M: One Million Image-Text Pairs for Histopathology}, 
      author={Wisdom Oluchi Ikezogwo and Mehmet Saygin Seyfioglu and Fatemeh Ghezloo and Dylan Stefan Chan Geva and Fatwir Sheikh Mohammed and Pavan Kumar Anand and Ranjay Krishna and Linda Shapiro},
      year={2023},
      eprint={2306.11207},
      archivePrefix={arXiv},
      primaryClass={cs.CV}
}

@misc{huang2025textdrivenmultiplanarvisualinteraction,
      title={Text-driven Multiplanar Visual Interaction for Semi-supervised Medical Image Segmentation}, 
      author={Kaiwen Huang and Yi Zhou and Huazhu Fu and Yizhe Zhang and Chen Gong and Tao Zhou},
      year={2025},
      eprint={2507.12382},
      archivePrefix={arXiv},
      primaryClass={cs.CV},
      url={https://arxiv.org/abs/2507.12382}, 
}
}

\end{document}